\documentclass[letterpaper, 10pt, conference]{ieeeconf}
\usepackage{times}
\usepackage[pdftex]{graphicx}

\usepackage{amsmath,amssymb,amsopn,amstext,amsfonts, amsthm}
\usepackage{cancel}
\usepackage[space]{cite}
\usepackage{pdfsync}
\usepackage{balance}
\usepackage{color}
\usepackage{mathtools}
\usepackage[ruled,norelsize, linesnumbered]{algorithm2e}
\usepackage{bm}
\usepackage{diagbox}
\usepackage{float}
\usepackage[outdir=./]{epstopdf}
\usepackage{url}
\usepackage{multirow}
\usepackage{makecell}
\usepackage{adjustbox}
\usepackage{tabularx, booktabs}
\let\labelindent\relax
\usepackage{enumitem}
\usepackage[font=small,skip=2pt]{caption}
\usepackage[linkcolor=black,citecolor=black,urlcolor=black,colorlinks=true]{hyperref}
\usepackage{subcaption}
\usepackage{bookmark}

\theoremstyle{remark}

\SetKw{Continue}{continue}
\makeatletter
\newcommand{\removelatexerror}{\let\@latex@error\@gobble}
\makeatother

\DeclareGraphicsExtensions{.pdf,.png,.jpg,.eps}
\IEEEoverridecommandlockouts
\title{\LARGE \bf  Predicting Vehicle Behaviors Over An Extended Horizon\\Using Behavior Interaction Network}
\author{Wenchao Ding$^\ast$, Jing Chen$^\dagger$, and Shaojie Shen$^\ast$%
\thanks{This work was supported by the Hong Kong PhD Fellowship Scheme.
$^\ast$These authors are with the HKUST-DJI Joint Innovation Laboratory, Department of Electronic and Computer Engineering, Hong Kong University of Science and Technology, Hong Kong, China.
$^\dagger$Jing Chen is with DJI Technology Co., Ltd.. Emails: {\tt\small wdingae@ust.hk, jing.chen@dji.com, eeshaojie@ust.hk}}%
}
\begin{document}

\maketitle
\thispagestyle{empty}
\pagestyle{empty}

\begin{abstract}
Anticipating possible behaviors of traffic participants is an essential capability of autonomous vehicles. Many behavior detection and maneuver recognition methods only have a very limited prediction horizon that leaves inadequate time and space for planning. To avoid unsatisfactory reactive decisions, it is essential to count long-term future rewards in planning, which requires extending the prediction horizon. In this paper, we uncover that clues to vehicle behaviors over an extended horizon can be found in vehicle interaction, which makes it possible to anticipate the likelihood of a certain behavior, even in the absence of any clear maneuver pattern. We adopt a recurrent neural network (RNN) for observation encoding, and based on that, we propose a novel vehicle behavior interaction network (VBIN) to capture the vehicle interaction from the hidden states and connection feature of each interaction pair. The output of our method is a probabilistic likelihood of multiple behavior classes, which matches the multimodal and uncertain nature of the distant future. A systematic comparison of our method against two state-of-the-art methods and another two baseline methods on a publicly available real highway dataset is provided, showing that our method has superior accuracy and advanced capability for interaction modeling.
\end{abstract}

\section{Introduction}\label{sec:introduction}
In recent years, there has been growing interest in building autonomous vehicles which can navigate naturally in complex environments. Despite the fact that perception techniques are maturing, autonomous vehicles are still criticized for being over conservative or socially incompliant. To make wise decisions, the planning module of autonomous vehicles needs to reason about long-term future outcomes, which requires predicting future behaviors in three to five seconds.

There is an extensive literature on the prediction of vehicles, which can be divided into two categories~\cite{hu2018simp}: \textbf{intention/maneuver/behavior} prediction and \textbf{motion/trajectory} prediction. The two categories have different outputs: the behavior prediction outputs high-level behaviors such as lane keeping (LK), lane change (LC), etc., while the motion prediction produces a time-profiled predicted trajectory. Recent works suggest that the motion prediction can be augmented by the behavior prediction to model the multimodal nature of future motion~\cite{hu2018simp,deo2018clstm,deo2018maneuvelmotion,ding2018onlinepred}. In this paper, we focus on the behavior prediction problem, and our method can be easily incorporated into motion prediction algorithms to generate time-profiled prediction.
\begin{figure}[t]
  \centering
  %left down right up
	\includegraphics[trim=0.0cm 1.6cm 0.0cm 1.0cm, width=0.48\textwidth]{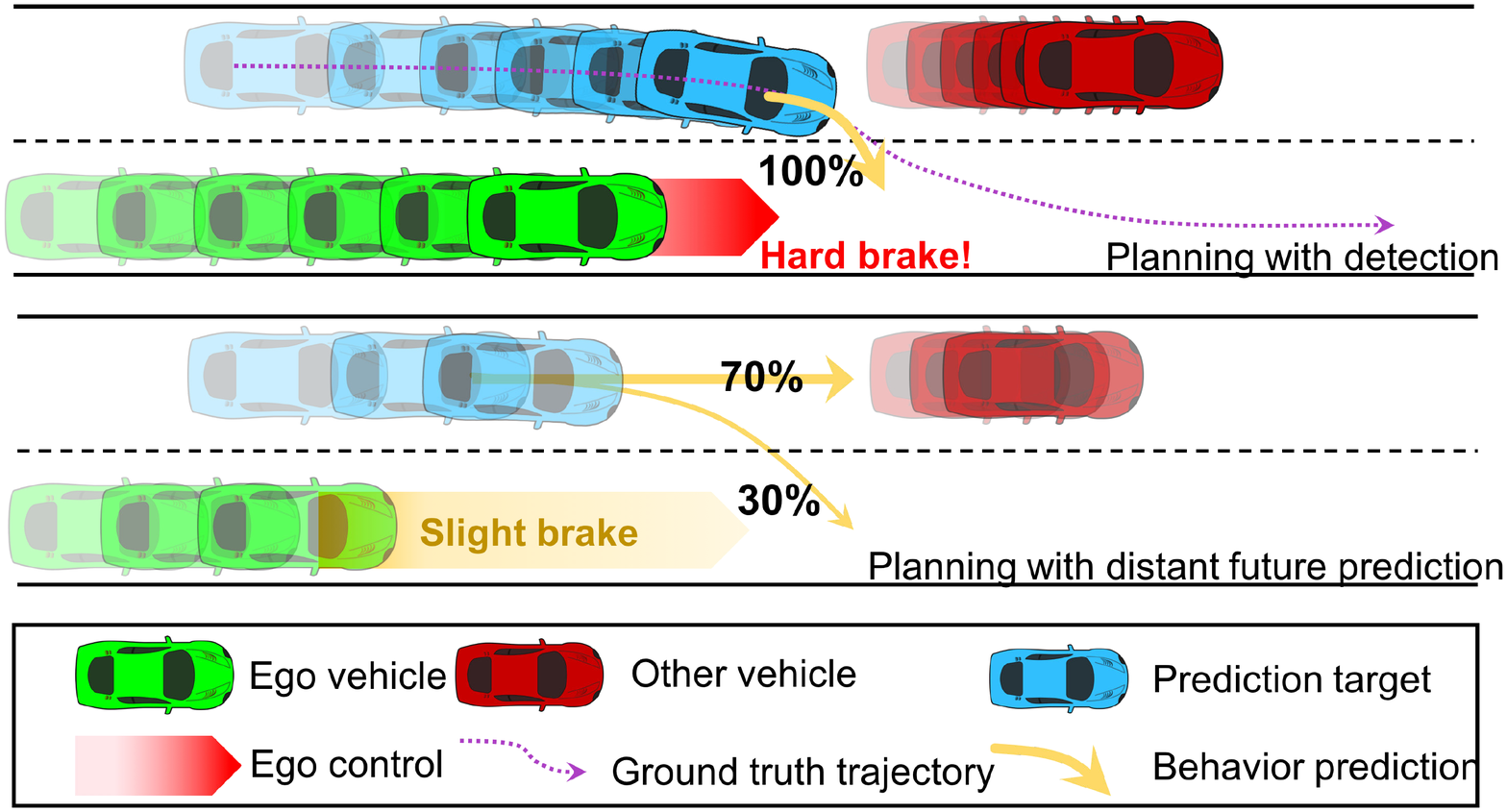}
  \caption{Illustration of the benefit of extending the prediction horizon. Assume that the \textbf{green} vehicle is the ego vehicle with the planning module, while the transparency of the vehicles represents the time elapsed. For a detection-based method, as shown on the top, the LC prediction is given when the \textbf{blue} vehicle has a clear LC pattern, which may result in a sudden braking of the ego vehicle due to the late discovery. However, from the interaction point of view, the \textbf{blue} vehicle is moving at a high speed and is blocked by the slowly moving \textbf{red} car. The \textbf{blue} vehicle has two interaction choices: brake to avoid collision or merge into the other lane. By learning from a large number of interaction patterns, the likelihood of LC can be estimated, even before the \textbf{blue} vehicle has a clear LC maneuver. More examples can be found in the video \url{https://www.youtube.com/watch?v=SuQzxAusU_0}.}\label{fig:motivation_example}
  \vspace{-0.6cm}
\end{figure}

An overview of prediction methods is provided in Sect.~\ref{sec:related_works}, where we find that many proposed methods are \textit{detection-based} and only have a limited prediction horizon. For example, a lane change is detected only $1.0$ s to $1.7$ s before the prediction target crosses the dividing line~\cite{houenou2013modelmaneuver,mandalia2005svm,schlechtriemen2014featureranking,woo2017detectiononprediction}. To avoid generating false alarms in the face of noisy maneuvers such as zigzagging, these methods tend to \textit{\textbf{detect} the behavior only when it is clearly happening}. Detection-based methods can achieve high accuracy and are suitable for a system like ADAS. However, the short prediction horizon is problematic for the planning module, as shown in Fig.~\ref{fig:motivation_example}.

To meet the needs of the planning module, the prediction is required to \textit{anticipate the likelihood of the behavior even before it is clearly happening}, and we call prediction of this kind prediction over an \textbf{extended horizon}. Such prediction typically requires reasoning about the behavior in three to five seconds. However, it is a challenging problem due to the increasing uncertainty with respect to the distance of the future. For example, at four seconds before a LC, the observed maneuver lacks features to make the LC distinguishable.

Obviously, additional information is required to reduce the uncertainty of the future. We uncover that social interaction is highly informative for prediction over the extended horizon. An illustrative example is shown in Fig.~\ref{fig:motivation_example}. In~\cite{alahi2016slstm}, Alahi~\textit{et al.} proposed a learning-based social pooling strategy for modeling pedestrian interactions, which became very popular and was later applied to vehicles~\cite{deo2018clstm}. However, using their methods agents appearing in the same spatial location will be weighted equally in spite of their different dynamics, which is problematic for a highly dynamic driving scenario, as shown in~Fig.~\ref{fig:occu_pooling}.

In this paper, we propose a novel vehicle behavior interaction network (VBIN) for modeling vehicle interactions. The novelty is that the VBIN is capable of \textit{learning to weight} the social effect of another agent on the prediction target based on their maneuver features and also relative dynamics (e.g., relative positions and velocities). Each interaction pair is dynamically weighted and then used to evaluate the total social effect on the prediction target. Contrary to the social pooling strategy, which is effective in modeling low-speed collision avoidance behaviors of pedestrians, our proposed method is suitable for highly dynamic driving scenarios where the dynamics of agents affect their importance in social interactions. Our method is end-to-end trainable.

We compare the proposed method with state-of-the-art interaction-aware models through systematic experiments on a publicly available real-world highway dataset, and show in Sect.~\ref{sec:experimental_results} that our novel VBIN structure can achieve significant improvements for behavior prediction over the extended horizon in terms of various critical metrics.

\section{Related Works}\label{sec:related_works}
\noindent\textbf{Detection and prediction.} Detection-based methods, such as maneuver recognition approaches, have been studied extensively. Houenou \textit{et al.}~\cite{houenou2013modelmaneuver} identify maneuver patterns such as LK and LC based on a handcrafted model which evaluates the similarity between the vehicle trajectory and lane center line. However, user-specified parameters and thresholds are required. Mandalia \textit{et al.}~\cite{mandalia2005svm} classify the LC maneuver using support vector machine (SVM), while Schlechitriemen \textit{et al.}~\cite{schlechtriemen2014featureranking} use Gaussian mixture models to estimate driving intentions. Woo \textit{et al.}~\cite{woo2017detectiononprediction} also adopt an SVM, and suggest using trajectory prediction to reject false alarms from the intention estimations. These methods detect LCs based on the target's own maneuver pattern and only have a limited prediction horizon. In contrast, our method captures the driver intention, even in the absence of any clear maneuver pattern, by making use of the vehicle behavior interactions.

\noindent\textbf{Behavior prediction and trajectory prediction.} Many methods focus on trajectory prediction and generate a time-profiled trajectory directly. Regression methods, such as linear regressions~\cite{nelder2004linearregression} and non-linear Gaussian process regression models~\cite{shalev2016planninggp,wang2008gphuman,quinonero2005sparsegp}, are extensively studied. Recurrent neural networks (RNNs) have recently been widely applied to various trajectory prediction tasks including trajectory prediction for pedestrians and vehicles. For vehicles, apart from predicting trajectories, many methods also aim to predict high-level intentions~\cite{woo2017detectiononprediction,schlechtriemen2014featureranking,gindele2015learningbehavior,lefevre2012detectintention}. Moreover, some recent works suggest that trajectory prediction can be augmented by the prediction of high-level semantic behaviors, e.g., LCs and insertions, due to the multimodal nature of the problem~\cite{deo2018clstm,hu2018simp,deo2018maneuvelmotion}. In this paper, we focus on the behavior prediction problem.

\begin{figure}[t]
  \centering
	\includegraphics[width=0.30\textwidth]{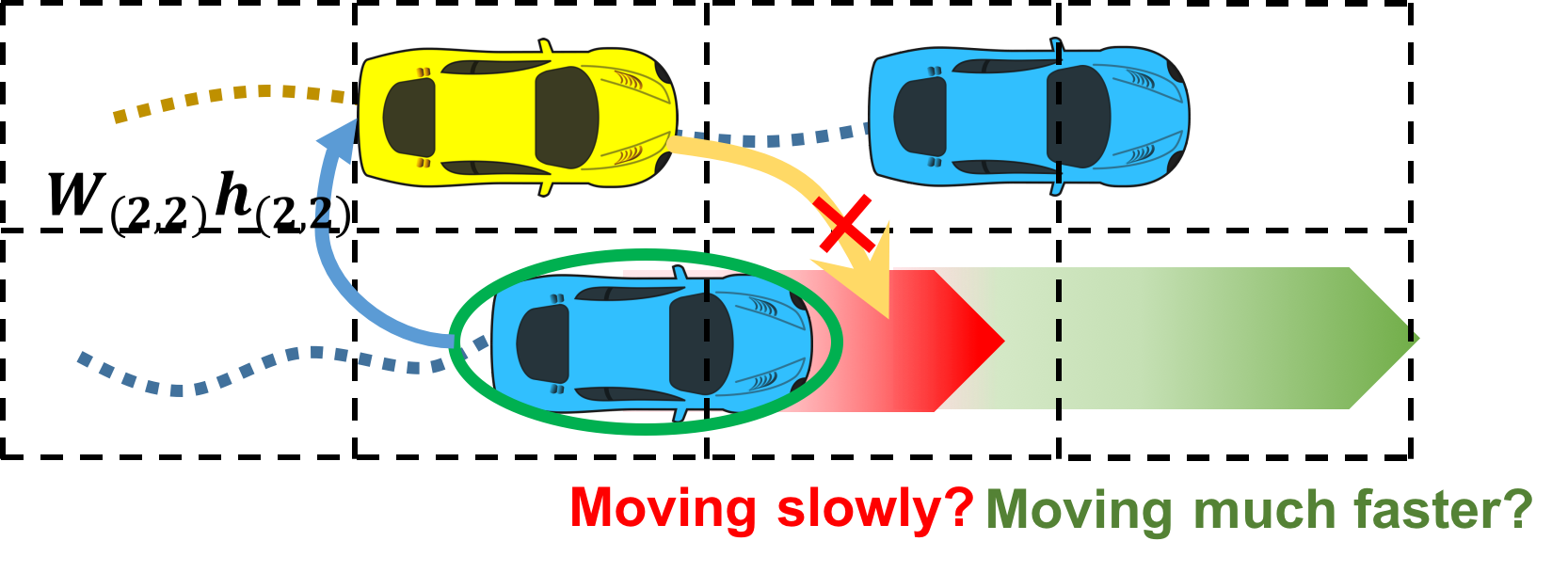}
	\caption{Illustration of the popular social pooling strategy~\cite{alahi2016slstm}. The RNN hidden states in the same spatial cell are pooled and passed to a fully connected layer to generate the total social effect on the target vehicle (\textbf{yellow}). Vehicles which have totally different dynamics in the same cell will share the same weight. However, as shown in the Illustration, if the vehicle that is highlighted with a circle is moving slowly, it is supposed to has a large weight since it blocks the LC route of the target vehicle, but if it is moving much faster than the target vehicle, it should has little impact on the target vehicle's LC. }\label{fig:occu_pooling}
  \vspace{-0.7cm}
\end{figure}

\noindent\textbf{Learning-based methods for various prediction tasks.}
Many learning-based prediction methods follow an RNN encoder-decoder structure~\cite{cho2014rnnecdc} and use it for predicting the trajectory of pedestrians~\cite{park2018s2s,lee2017desire,zou2018rnnmdp} and vehicles~\cite{deo2018clstm,lee2017desire}. RNNs are an alternative to the traditional methods (e.g., SVM, Gaussian process)~\cite{zyner2017lstmgps,dang2017time} for capturing maneuver patterns. Some works move one step further and consider the multi-agent interactions among pedestrians~\cite{alahi2016slstm,lee2017desire,bartoli2017contextlstm,varshneya2017slstmspatialcontext} and vehicles~\cite{lee2017desire,deo2018clstm} in the RNN structure. The interaction-aware models for vehicles mainly adopt a social pooling strategy using RNN hidden states, which may be problematic for a highly dynamic scenario, as elaborated in the following.

\noindent\textbf{Modeling interactions.}
Modeling social interaction in RNNs started with the seminal work by Alahi \textit{et al.}~\cite{alahi2016slstm}, which proposes a social occupancy grid pooling mechanism, as shown in Fig.~\ref{fig:occu_pooling}. This design is very effective for pedestrian trajectory prediction in a crowd~\cite{lee2017desire,bartoli2017contextlstm,xue2018ss}, and is suitable for unstructured environments and modeling collision avoidance behaviors. It is directly applied to vehicle prediction in~\cite{lee2017desire} and~\cite{deo2018clstm}. However, for highly dynamic on-road driving, spatial locations should not be the only reference for weighting social effects, as shown in Fig.~\ref{fig:occu_pooling}. Going beyond these methods, our proposed VBIN can learn to weight the interactions automatically based on the relative dynamics of both agents.

\section{Methodology}\label{sec:methodology}
\subsection{Problem Formulation}\label{subsec:problem_formulation}
The motivation behind our proposed behavior prediction method is to facilitate the development of socially compliant planning algorithms. To this end, we describe the problem from the planning perspective to make clear what requirements the planning module for the prediction. We follow multi-policy decision making (MPDM)~\cite{cunningham2015mpdm,galceran2015mpdmchangept} for a theoretically-grounded multi-agent planning formulation.
Note that our proposed method is interaction-aware and can serve as a further developed prediction module for this planning framework.

At time t, a vehicle $v$ can take an action $a_t^{v}\in \mathcal{A}^{v}$ to transition from a state $x_t^{v}\in \mathcal{X}^v$ to $x_{t+1}^v$ and make an observation $z_t^v\in \mathcal{Z}^v$. For example, a state $x_t^{v}$ can be a tuple of the pose, velocity and acceleration, an action $a_t^{v}$ can be a tuple of the controls for steering, throttle, braking, etc., and an observation $z_t^v$ can be a tuple of estimated poses and velocities of other vehicles. Let $e$ denote the ego (i.e., controlled) vehicle and $V_e$ denote the set of vehicles in a local neighborhood of the ego vehicle $e$. Let $z_t^{e,v}$ denote the observation made by the ego vehicle about its nearby vehicle $v\in V_e$, let $x_t$ include all state variables $x_t^{v}$ for all vehicles at time $t$, let $a_t\in\mathcal{A}$ be the actions of all vehicles, and let $z_t \in \mathcal{Z}$ be the observations made by all vehicles. From the planning perspective, the goal is to find an optimal policy $\pi^{\prime}$ that maximizes the expected reward over a given decision horizon $H$, where a policy is a mapping $\pi:\mathcal{X}\times \mathcal{Z}^{v} \rightarrow \mathcal{A}^{v}$. $\pi^{\prime}$ can be obtained by solving a partially observable Markov decision process (POMDP) with the following probability transition:
\vspace{-0.1cm}
\begin{equation}
p(x_{t+1})= \prod_{v\in V}\!\!\underset{\mathcal{X}^v\mathcal{Z}^{v}\mathcal{A}^{v}}{\int\!\!\!\!\int\!\!\!\!\int}\!\!p^v(x_t^v, x_{t+1}^v,z_t^v,a_t^v)da_t^v dz_t^v dx_t^v,
\end{equation}
where $p^v(x_t^v, x_{t+1}^v,z_t^v,a_t^v)$ denotes the joint density for a single vehicle.
The MPDM make the assumption on the driver model $p(a_t^v|x_t^v,z_t^v)$ that the agents are executing a policy/behavior from a discrete set of behaviors, such as LC and LK. Mathematically, the assumption is
\vspace{-0.2cm}
\begin{equation}
p(a_t^v|x_t^v,z_t^v)=p(a_t^v|x_t,z_t^v,\pi_t^v)\underbrace{p(\pi_t^v|x_t^v, z_t^v)}_\text{behavior decision},
\end{equation}
where $\pi_t^v$ belongs to the discrete set of behaviors $\Pi$. The ego vehicle needs to infer $p(\pi_t^v|x_t^v,z_t^v)$ for all $v\in V_e$ via $p(\pi_t^v|\mathbf{z}^e_{0:t})$, where $\mathbf{z}^e_{0:t}$ is a time series of observations of the nearby vehicles.

Many prediction methods actually simplify the problem by further approximating $p(\pi_t^v|\mathbf{z}^e_{0:t})$ using $p(\pi_t^v|\mathbf{z}^{e,v}_{0:t})$, where $p(\pi_t^v|\mathbf{z}^{e,v}_{0:t})$ denotes the observations of vehicle $v$ only. Specifically, the predicted behavior is only based on past observations of each vehicle itself. The approximation ignores the potential interactions and results in a short prediction horizon, as shown in Fig.~\ref{fig:motivation_example}. It also becomes clear that learning-based methods try to find a parametric representation of the policy likelihood $p(\pi_t^v|\mathbf{z}^e_{0:t},\bm{\theta})$ using the neural network parameters $\bm{\theta}$. The idea of the proposed method is to uncover the potential interactions lying inside $\mathbf{z}^e_{0:t}$ through the specially designed VBIN structure and yield a better parametric estimation of $p(\pi_t^v|\mathbf{z}^e_{0:t})$ which holds for a long decision horizon $H$.
\begin{figure}[t]
  \centering
	\includegraphics[width=0.35\textwidth]{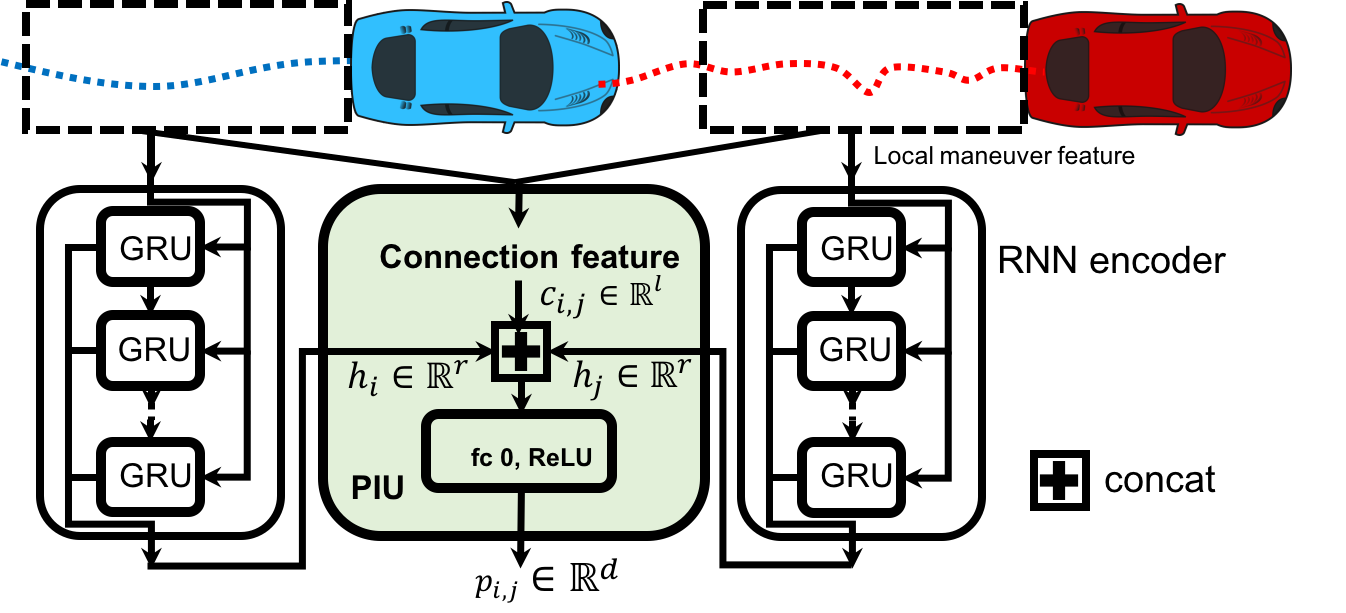}
	\caption{VBIN building block: pairwise interaction unit (PIU). RNNs are implemented using gated recurrent units (GRUs)~\cite{cho2014rnnecdc}, and the hidden states are used as the input of the PIU. }\label{fig:piu}
  \vspace{-0.7cm}
\end{figure}

\subsection{RNN Encoder Network}\label{subsec:rnn_encoder}
As introduced in Sect.~\ref{subsec:problem_formulation}, the network is supposed to be amenable to encoding the observed states of all vehicles. To accomplish this task, we start with obtaining the encoding for each individual agents, which provides a compact representation of the vehicle's maneuver. We adopt the popular RNN encoder network~\cite{cho2014rnnecdc} to obtain the maneuver encoding for each individual vehicle. We first extract features from a time series of the observed states of a given vehicle via a feature selection function $f_m(\mathbf{z}^{e,v}_{0:t})$. The selection of features depends on the design of the behavior class, and in Sect.~\ref{subsec:implementation_details}, we provide a practical example of the feature selection for LC behavior.

For brevity, we do not include details of the RNN encoder structure, and refer interested readers to~\cite{alahi2016slstm} and \cite{lee2017desire} for the details. It is worth noting that, during the training phase, all the RNNs share the same set of weights. After the RNN encoding, the maneuver history of each vehicle $v\in V_e$ is encoded in a vector $h_v \triangleq \text{RNN}(f_m(\mathbf{z}^{e,v}_{0:t}))\in \mathbb{R}^{r}$, where $r$ denotes the size of the RNN encoding. All the encodings can be computed and stored before the inference of the VBIN.

\subsection{Vehicle Behavior Interaction Network (VBIN)}\label{subsec:VBIN}
\noindent \textbf{Pairwise interaction unit (PIU).} The basic element of the VBIN is the pairwise interaction unit (PIU). The PIU \textit{learns to weight} the social effect of an interaction pair based on their maneuver histories and relative dynamics. As shown in Fig.~\ref{fig:piu}, the PIU takes three inputs: two RNN encodings $h_i$ and $h_j$ of a pair of vehicles, with $i\in V_e$, $j\in V_e$ and $i\neq j$, and the connection feature $c_{i,j}$. The connection feature $c_{i,j}$ is extracted via another feature extraction function, $c_{i,j} = f_c(\mathbf{z}^{e,i}_{0:t}, \mathbf{z}^{e,j}_{0:t}) \in \mathbb{R}^l$, which is based on the history of both vehicles and represents the relative states of both vehicles. For example, $c_{i,j}$ can include the relative positions and relative velocities of the two vehicles.

In other words, the hidden states $h_i$ and $h_j$ are responsible for describing the maneuver in each vehicle's \textit{local reference frame}, and are invariant no matter which prediction target it is. The connection feature $c_{i,j}\in\mathbb{R}^{l}$ describes the ``translation'' of two reference frames and the dynamics of both vehicles. The PIU is dedicated to ``weighting'' the two hidden states given the additional information of translation and vehicle dynamics. Unlike the conventional social pooling~\cite{alahi2016slstm}, which weights the hidden states purely based on spatial location, the PIU learns different weighting strategies for different relative states of each interaction pair. The output of the PIU unit is denoted as $p_{i,j} \triangleq \text{PIU}(h_{i}, h_{j}, c_{i,j}) \in \mathbb{R}^d$, with $d$ denoting the size of the PIU embedding.

\noindent \textbf{Neighborhood interaction unit (NIU).}
Before stepping into the details of the NIU, we give a formal definition of ``neighborhood''. In the series of works on pedestrian trajectory prediction~\cite{alahi2016slstm},~\cite{deo2018clstm} and~\cite{bartoli2017contextlstm}, the ``neighborhood'' is defined by a grid centered at the prediction target and each vehicle is associated with one cell. However, vehicles travel in a semi-structured environment where there are semantic elements, such as lanes, which makes the occupancy grid not the best choice. \cite{hu2018simp} suggests selecting a fixed number of reference vehicles around the prediction target based on the environment semantics such as lanes. Taking the highway scenario for example, vehicles tend to interact with the nearest vehicles in the current and neighboring lanes, and these neighboring vehicles will be informative for interaction modeling~\cite{hu2018simp}.

\begin{figure}[t]
  \centering
	\includegraphics[trim=1.5cm 4.7cm 1.5cm 4.6cm, width=0.37\textwidth]{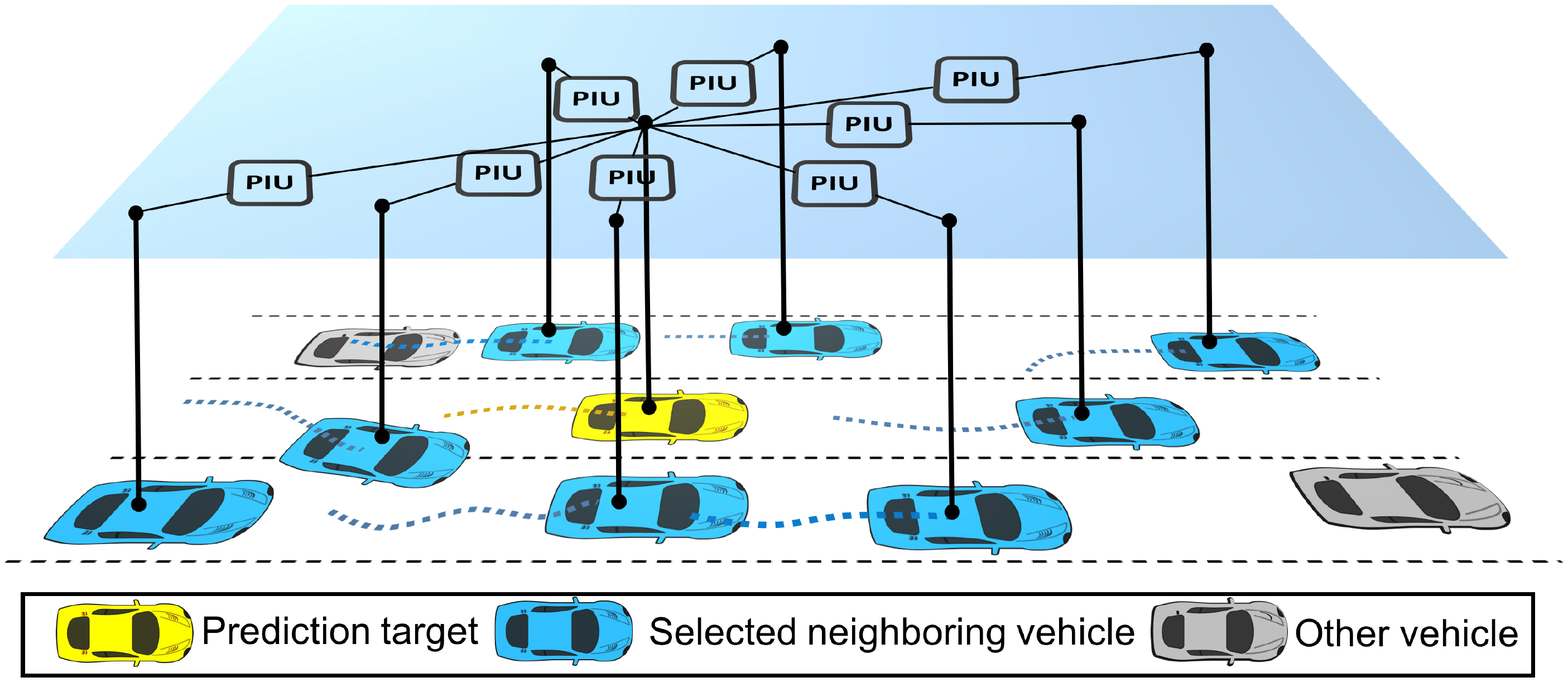}
  \caption{Illustration of the PIU connections for each selected neighboring vehicle. }\label{fig:reference_vehicle}
  \vspace{-0.3cm}
\end{figure}

In this section, we provide a selection strategy for a highway scenario as an example, as shown in Fig.~\ref{fig:reference_vehicle}. The selection process is as follows: 1) Select the two vehicles at the front and rear in the current lane. 2) Select the two vehicles with the closest longitudinal coordinates to the target vehicle in the neighboring lanes. 3) Select the vehicles immediately at the front and the vehicle immediately at the rear w.r.t. the neighboring vehicle. In the case of the non-existence of such neighboring vehicles, we create a virtual vehicle with the same longitudinal velocity as the target vehicle and a far longitudinal relative distance (e.g., $100$ m) in the same direction. Note that the neighboring vehicles are determined online. For the consecutive rounds of predictions, the neighboring vehicles may vary but the prediction results are still consistent, as shown in the video.

\begin{figure}[t]
  \centering
	\includegraphics[width=0.35\textwidth]{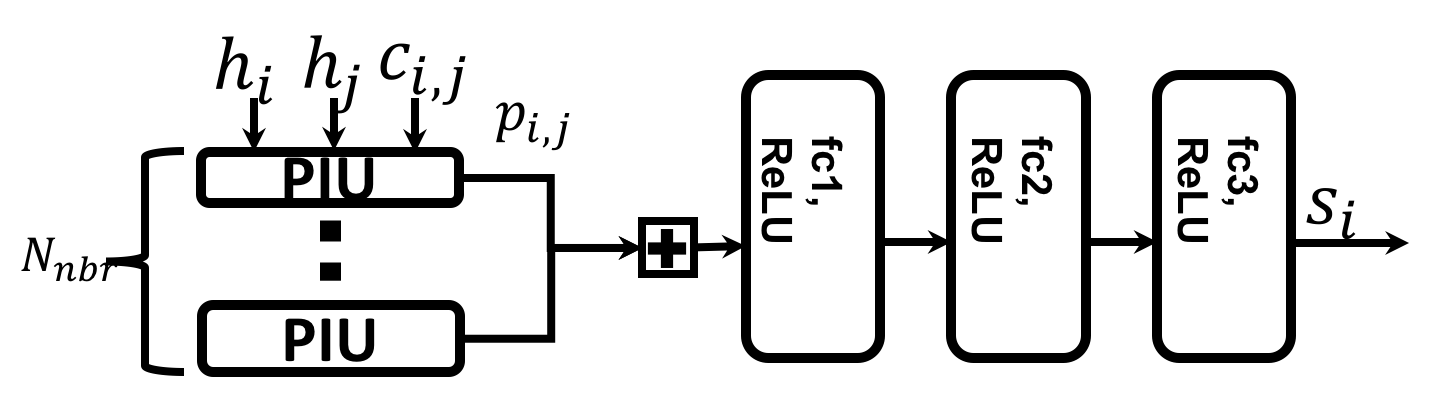}
  \caption{VBIN building block: neighborhood interaction unit (NIU). All the PIUs are identical.}\label{fig:niu}
  \vspace{-1.5cm}
\end{figure}

Let $\mathcal{N}_i$ be the set of selected neighboring vehicles around the target vehicle $i\in V_e$. We denote the number of selected neighboring vehicles as $N_{\text{nbr}}=|\mathcal{N}_i|$. Like~\cite{hu2018simp}, we use a fixed $N_{\text{nbr}}$ for each prediction target. As shown in Fig.~\ref{fig:reference_vehicle}, we establish the PIU link between the prediction target $i$ and each selected vehicle $j\in \mathcal{N}_i$. The $N_{\text{nbr}}$ PIU embeddings are concatenated and passed to three cascaded fully connected layers with ReLU activation. The output of the NIU unit is denoted as $s_{i} \in \mathbb{R}^{n}$, with $n$ denoting the NIU embedding size. At a high level, $s_{i}$ represents the total social effect of the neighboring vehicles on the prediction target $i$.

\begin{figure}[t]
  \centering
	\includegraphics[width=0.386\textwidth]{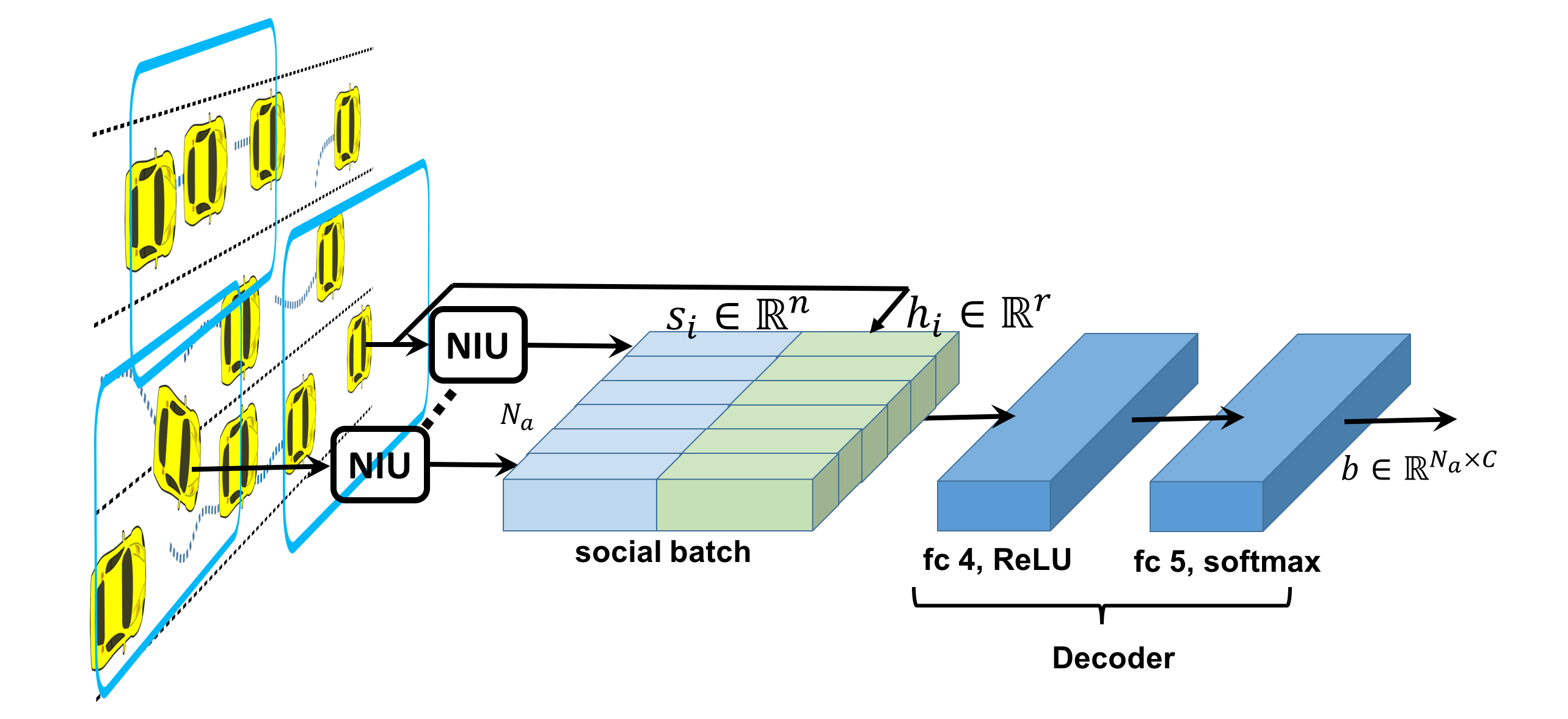}
  \caption{Illustration of the VBIN structure. All the NIUs are identical.}\label{fig:model_integration}
  \vspace{-1.0cm}
\end{figure}

\noindent \textbf{Behavior decoding.}
As mentioned above, $s_{i}$ represents the social effect applied to vehicle $i$. We then concatenate $s_{i}$ with its original local maneuver encoding $h_i$. The concatenated vector now contains both the features extracted from its own maneuver history and the features extracted from the neighborhood interaction. Applying the above process to a total number of $N_a$ prediction targets, we obtain a social batch, with each row representing the combined encoding for each individual vehicle, as shown in Fig.~\ref{fig:model_integration}. The social batch is passed to cascaded fully connected layers with ReLU activation and another softmax layer to decode the output, which is the likelihood $b\in\mathbb{R}^{N_a\times C}$ of $C$ behavior classes for each predicted vehicle.

\noindent \textbf{Scalable deployment.}
In the following, we show that the inference time of the VBIN does not scale with $N_a$ and is amenable to scalable deployment. Specifically, since the PIUs and NIUs are all identical, we can precompute and reuse the following tensors: 1) precompute the connection features for the neighborhood of each agent, which forms a tensor of size $N_a \times N_{\text{nbr}} \times l$, and store an index tensor of size $N_a \times N_{\text{nbr}} \times 1$, which records the index of the corresponding neighboring vehicle; 2) use the RNN to encode $N_a$ local maneuver features and get the resulting hidden tensor of size $N_a \times r$.
\begin{figure}[htb]
  \centering
	\includegraphics[width=0.43\textwidth]{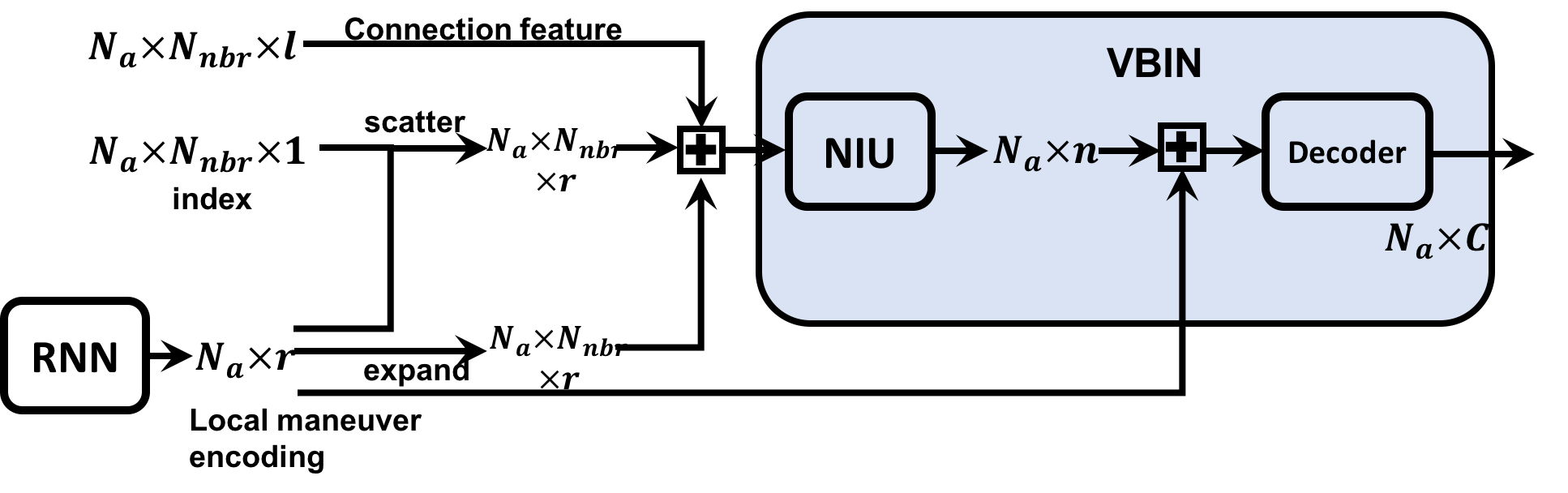}
  \caption{Illustration of the dataflow of the system.}\label{fig:dataflow}
  \vspace{-0.3cm}
\end{figure}
After the preparations, we can run a forward pass, as shown in Fig.~\ref{fig:dataflow}. The input tensor of the NIU is of size $N_a \times N_{\text{nbr}} \times (2r+l)$, and it combines each pair of hidden states and the corresponding connection feature. Inside the NIU, the PIU resizes and embeds the input tensor to size $N_a \times N_{\text{nbr}} \times d$, and its second and third dimension will be further flattened. After the output layers of the NIU, the tensor is of size $N_a \times n$, and it will be concatenated with the original hidden states and decoded, as shown in Fig.~\ref{fig:model_integration}. $N_a$ can be regarded as the batch size and will not affect the inference time.

\section{Experimental Results}\label{sec:experimental_results}
We apply our proposed method to a highway scenario to validate the performance and illustrate the implementation details for highway application.

\subsection{Dataset}\label{subsec:dataset}
We use a real traffic dataset, NGSIM, which is publicly available online~\cite{ngsim}. We use data collected from highways US-101 and I-80. The dataset consists of trajectories of real traffic captured at 10 Hz, and also provides mild, moderate and congested traffic conditions. We define the behavior classes as \textit{lane keeping}, \textit{lane change left} and \textit{lane change right}. We extract $1,669$ LCs (and LKs) from the dataset and a total of $509$ LCs (and LKs) are used for testing. The training and testing set are totally disjoint.

\subsection{Baselines}\label{subsec:baselines}
We compare our method with the following methods:
\begin{itemize}[leftmargin=*]
  \item \textit{Vanilla LSTM} (\textbf{VLSTM}). This is a simplified setting of our model, where we remove the interaction network VBIN.
  \item \textit{Social LSTM} (\textbf{SLSTM}). This is from the seminal work~\cite{alahi2016slstm} on interaction-aware trajectory prediction, and we change the output to the likelihood of behavior classes.
  \item \textit{Convolutional social pooling LSTM} (\textbf{CLSTM}). This is a modification of SLSTM~\cite{deo2018clstm}. Both SLSTM and CLSTM use the social pooling strategy. The difference is that CLSTM uses convolutional layers for the pooled hidden states.
  \item \textit{Semantic-based intention and motion prediction} (\textbf{SIMP}). We use the implementation from～\cite{hu2018simp} and only adopt the intention prediction part.
\end{itemize}

CLSTM and SIMP are two state-of-the-art methods, originally verified on the NGSIM dataset. We adopt the publicly available implementation~\cite{deo2018clstmgithub} of CLSTM and SLSTM, and train the model on our generated dataset. Since the source code of~\cite{hu2018simp} is not officially available, we implement it according to the implementation details provided in~\cite{hu2018simp}.

\subsection{Implementation Details}\label{subsec:implementation_details}
The output of the feature selection function $f_m(\mathbf{z}^{e,v}_{0:t})$ is shown in Tab.~\ref{tab:local_maneuver_feature}.
\begin{table}[h]
  %Data used: 090423, relaxed criterion, only critical fns, only use 0-79
  \centering
	\caption{Feature selection for local maneuver encoding.\label{tab:local_maneuver_feature}}
  \resizebox{0.48\textwidth}{!}{
  \begin{tabular}{@{}ll@{}}
	\toprule
	Feature & Description \\
  \midrule
  $x^{\text{lat}}$ &  Lateral coordinate of the center of the vehicle\\
  $x^{\text{long}}$ & longitudinal coordinate of the center of the vehicle\\
  $d^{\text{lat}}_{\text{clc}}$ & Lateral distance to the current lane center normalized by lane width  \\
  $v^{\text{long}}$ & longitudinal velocity in the lane reference frame\\
  $v^{\text{lat}}$ &  Lateral velocity in the lane reference frame\\
  $\theta$ & Vehicle orientation with respect to the lane longitudinal direction\\
  \bottomrule
  \end{tabular}}
  \vspace{-0.2cm}
\end{table}
The length of the observation window is set to $2$ s (20 frames). For the sequence of local maneuver features to be fed to the RNN, we subtract the $x^{\text{lat}}$, $x^{\text{long}}$ of the last frame of the sequence to remove the absolute translation. The length of the prediction window is set to $4$s (40 frames).
The criterion for labeling LC is that in the prediction window the target vehicle crosses the lane dividing line.
The labeling process is conducted in a sliding window manner starting from $8$ s before the LC.
All the methods output the likelihood of the three behavior classes, and the prediction decision is made by taking the class with the maximum likelihood. The structure of the RNN encoder is the same as~\cite{lee2017desire} with 128 hidden states. The loss function is the negative log-likelihood (NLL) loss with three classes. All the models are implemented using Pytorch~\cite{paszke2017automatic}.

The connection feature consists of six elements: relative longitudinal and lateral distances, longitudinal and lateral velocities of the target vehicle, and longitudinal and lateral velocities of the neighboring vehicle. The detailed network specifications are shown in Tab.~\ref{tab:network_spec}.
\begin{table}[h]
  % \vspace{-0.3cm}
  \centering
	\caption{Network specifications of the VBIN.\label{tab:network_spec}}
  \resizebox{0.45\textwidth}{!}{
  \begin{tabular}{@{\extracolsep{6pt}}cccccccccc@{}}
  \toprule
  \multicolumn{4}{c}{\textbf{PIU}} & \multicolumn{3}{c}{\textbf{NIU}} & \multicolumn{3}{c}{\textbf{Decoder}}\\
  \cline{1-4} \cline{5-7} \cline{8-10}
  $r$ & $l$ & $d$ & fc0 I/O & fc1 I/O & fc2 I/O & fc3 I/O& $n$ & fc4 I/O & fc5 I/O\\
	\midrule
  48  & 6   & 64  & 102/64  & 512/400 & 400/400 & 400/48 & 48  & 96/48   & 48/3\\
  \bottomrule
  \end{tabular}}
  \vspace{-0.3cm}
\end{table}

\subsection{Evaluation Metrics}\label{subsec:evaluation_metrics}
\noindent \textbf{Accuracy.} Precision, recall and F1-score are the three representative metrics quantifying the accuracy for classification problem. In the case of LC prediction, we define two positive classes (lane change left and right) and one negative class (LK), similar to~\cite{hu2018simp} and~\cite{dang2017time}. However, we argue that for different times-to-lane-change (TTLCs), there is particular emphasis on different errors. We define the the false negatives (FNs) inside the duration of TTLCs less than $1.5$ s as \textit{critical FNs}, since it is too close to the actual LC and any mis-detection will be extremely dangerous. Note that the criterion is more strict than that adopted by~\cite{woo2017detectiononprediction}, where the FN is tolerated as long as the vehicle hasn't crossed the dividing line. We define the false positives (FPs) inside the duration of TTLCs larger than $5.5$ s as \textit{critical FPs}, since a lane change generally takes $3.0$ to $5.0$ s～\cite{toledo2007modelinglanechange} and a too early LC alarm will cause disturbance.

\noindent \textbf{Negative log-likelihood (NLL).} Apart from the traditional accuracy metrics, we adopt the NLL loss of multi-class classification to measure the classification uncertainty. Uncertainty constantly decreases as the TTLC approaches $0$, but different methods perform differently for uncertainty reduction. Note that the evaluation of NLL loss is irrelevant to the definition of positive classes.

\noindent \textbf{Average prediction time.} Prediction time represents the average \textit{effective} prediction horizon for LCs. Generally, the prediction time is obtained by recording the time of the first successful LC prediction. However, we consider a strict strategy and only record the starting point of a series of \textit{consistent} LC predictions, where two consecutive LC predictions should not contain a gap larger than three frames.

\subsection{Analysis}\label{subsec:analysis}
\noindent \textbf{Overall accuracy.} In Tab.~\ref{tab:compare_accuracy}, we report the precision, recall and F1-score of all the methods. The calculation of the recall is based on the critical FNs representing the most unacceptable misclassifications. Note that we start the evaluation from an early phase, namely, $8$ s before the target vehicle crosses the lane dividing line, which may introduce more FPs. Only the frames which contain enough ground truth observation and prediction are picked. There are a total of $13,338$ frames of prediction.
\begin{table}[b]
  %Data used: 090423, relaxed criterion, only critical fns
	\centering
	\caption{A comparison of different prediction methods. The methods marked with $\dagger$ are interaction aware.\label{tab:compare_accuracy}}
	\resizebox{0.45\textwidth}{!}{
  \begin{tabular}{@{}lcccc@{}}
	\toprule
	 \textbf{Method}
	&\textbf{Precision}
	&\textbf{Recall}
	&\textbf{F1-score}
	&\textbf{Ave. Predict Time (s)}\\
	\midrule
	$\text{VLSTM}$ & 0.802 & 0.920 & 0.857 & 1.999\\
	\hline
	$\text{SLSTM}^\dagger$\cite{alahi2016slstm}  & 0.890 & 0.900 & 0.895  & 2.355\\
	$\text{CLSTM}^\dagger$\cite{deo2018clstm}    & 0.885 & 0.932 & 0.908  & 2.294\\
	$\text{SIMP}^\dagger$~\cite{hu2018simp}      & 0.873 & 0.931 & 0.901  & 2.102\\
  \hline
	$\text{Our VBIN}^\dagger$ & \textbf{0.922} & \textbf{0.967} & \textbf{0.944} & \textbf{2.622}\\
  \bottomrule
  \end{tabular}}
  \vspace{-0.3cm}
\end{table}

\begin{figure*}[t]
	\centering
		%trim left-down-right-up
		\begin{subfigure}[b]{0.33\textwidth}
			\includegraphics[trim=0.0cm 0.0cm 0.0cm 0.0cm, width =\textwidth]{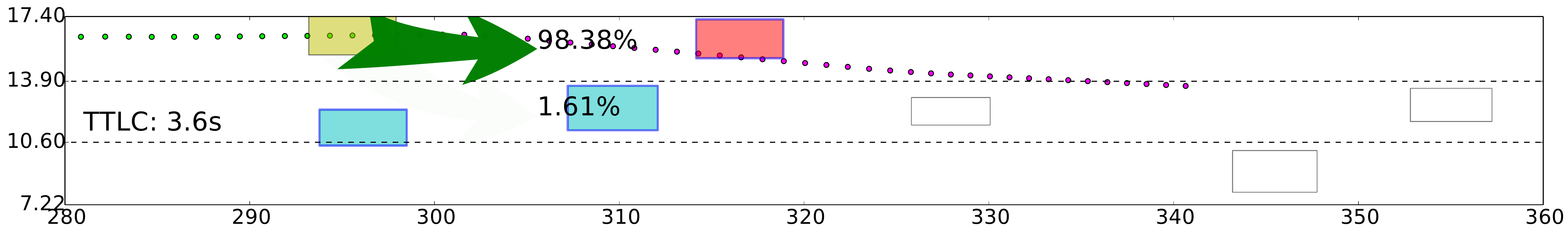}
			\includegraphics[width =\textwidth]{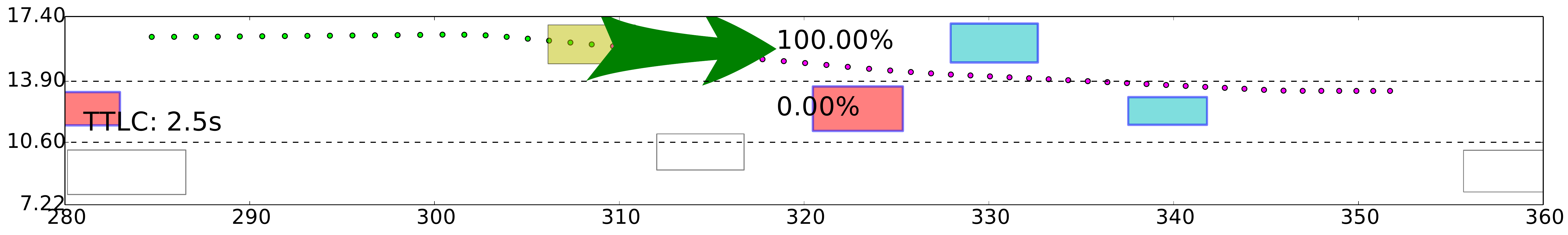}
			\includegraphics[width =\textwidth]{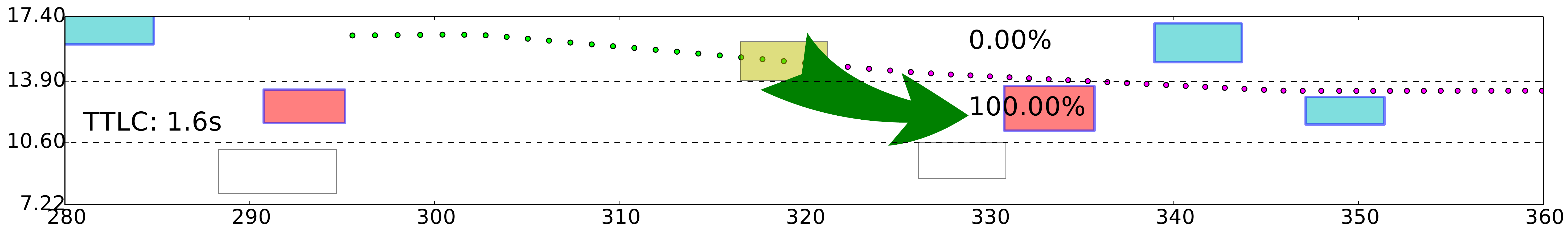}
			\caption{Baseline: VLSTM}\label{fig:comparison_vlstm}
		\end{subfigure}%
		\begin{subfigure}[b]{0.33\textwidth}
			\includegraphics[trim=0.0cm 0.0cm 0.0cm 0.0cm, width =\textwidth]{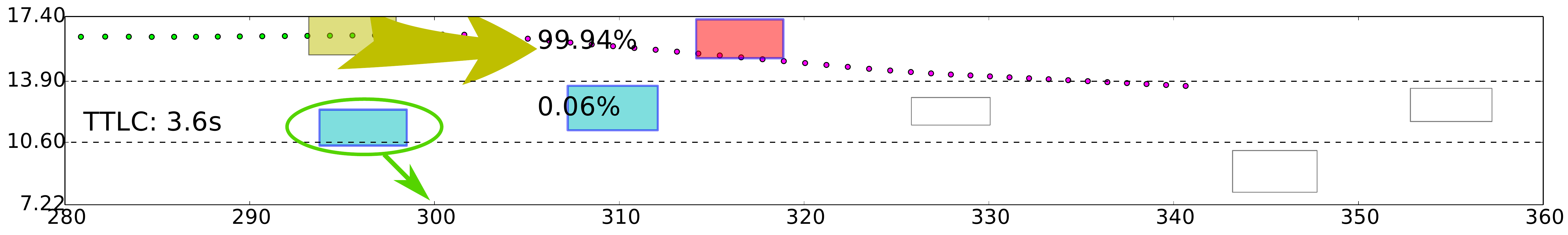}
			\includegraphics[width =\textwidth]{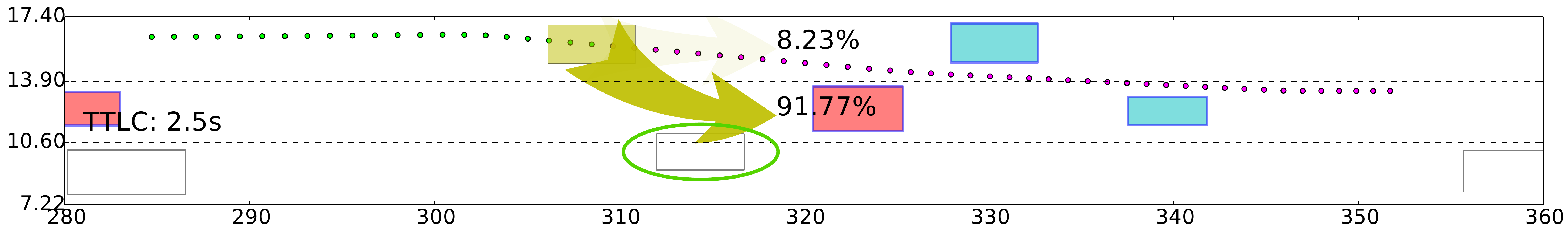}
			\includegraphics[width =\textwidth]{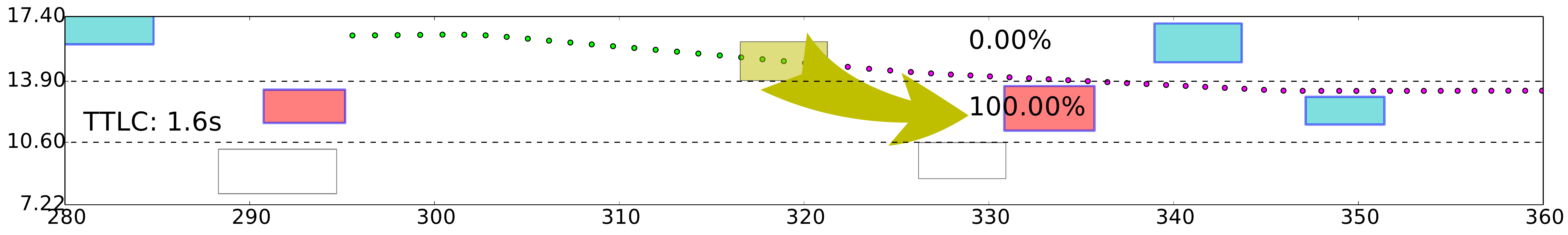}
			\caption{Baseline: CLSTM}\label{fig:comparison_clstm}
		\end{subfigure}%
		\begin{subfigure}[b]{0.33\textwidth}
			\includegraphics[trim=0.0cm 0.0cm 0.0cm 0.0cm, width =\textwidth]{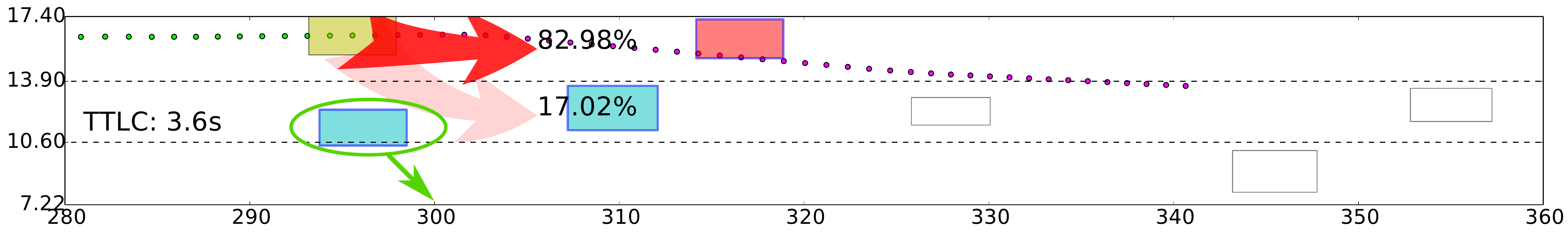}
			\includegraphics[width =\textwidth]{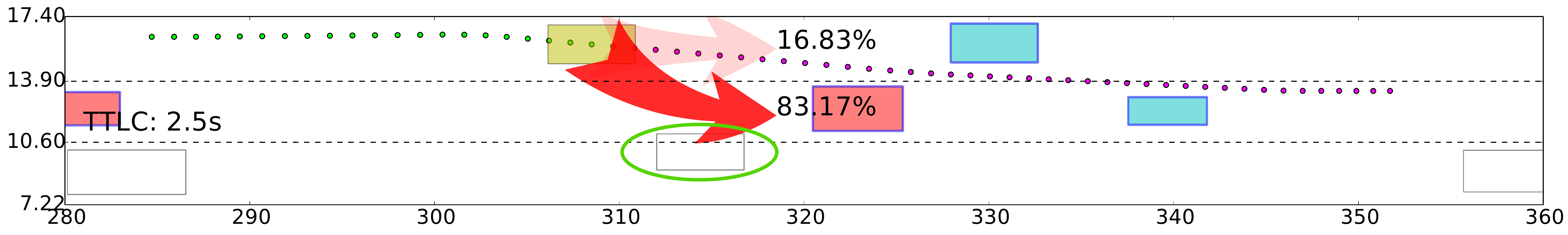}
			\includegraphics[width =\textwidth]{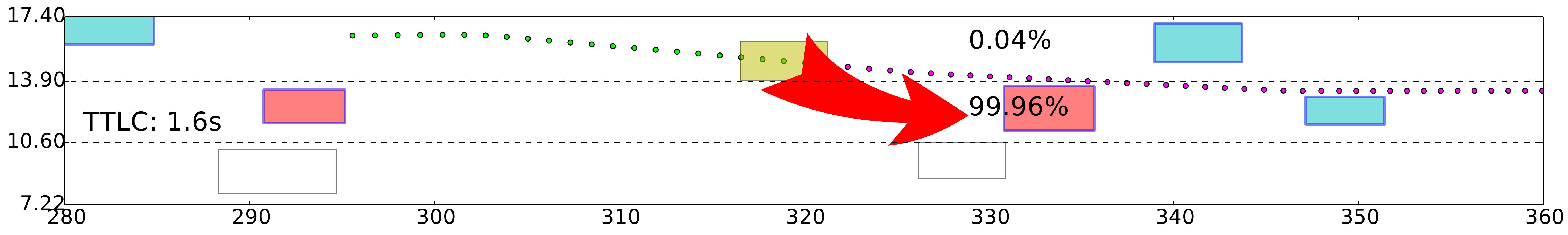}
			\caption{Our VBIN}\label{fig:comparison_vbin}
		\end{subfigure}%
  \caption{Illustration of the comparison on the NGSIM dataset. The vehicle marked in \textbf{yellow} is the prediction target. The vehicles filled with \textbf{cyan} and \textbf{red} are the selected neighboring vehicles. The pair of each \textbf{red} vehicle and target vehicle may have potential interaction according to the RSS safety model~\cite{shalev2017rss}. Note that we do not use any handcrafted models and the RSS is used to aid understanding of the scenario. The behavior predictions are marked in arrows, and the transparency represents the magnitude of the likelihood. }\label{fig:comparison_tuple}
	\vspace{-0.3cm}
\end{figure*}

According to Tab.~\ref{tab:compare_accuracy}, the proposed VBIN has the highest precision, recall and F1-score among all the baseline methods. Note that the precision of SIMP reported in Tab.~\ref{fig:comparison_tuple} is slightly lower than that in~\cite{hu2018simp}, since~\cite{hu2018simp} does not include the FPs from $8$ s to $4$ s. Compared to CLSTM, our method achieves a higher recall of $0.967$. Our conjecture is that CLSTM may underestimate the likelihood of LC due to the social pooling strategy, especially when the neighboring lane is congested, which may result in more FNs.

The critical FPs and FNs are also listed in Tab.~\ref{tab:critical_classfication_err}. Our proposed method has the least critical FNs, at $206$, which indicates that our method preserves the accuracy of the near future. Also, our proposed method has the least critical FPs, at $322$, which means that the interaction information is useful for rejecting false alarms. Compared to～\cite{woo2017detectiononprediction}, which uses trajectory prediction on an artificial potential field to reject FPs, our proposed method is completely data driven and free of hand tuning.

\begin{table}[h]
  %Data used: 090423, relaxed criterion, only critical fns, only use 0-79
  \centering
	\caption{Critical misclassifications over $13,338$ predictions.\label{tab:critical_classfication_err}}
  \resizebox{0.45\textwidth}{!}{
  \begin{tabular}{@{}lccccc@{}}
	\toprule
	\textbf{Method} &\text{VLSTM} & $\text{SLSTM}^\dagger$ & $\text{CLSTM}^\dagger$ & $\text{SIMP}^\dagger$ & $\text{Our VBIN}^\dagger$ \\
	\midrule
	\textbf{Critical FNs} $(<1.5s)$   & 435 & 597 & 400 & 477 & \textbf{206}\\
	\textbf{Critical FPs} $(>5.5s)$ & 564 & 349 & 387 & 562 & \textbf{322}\\
  \bottomrule
  \end{tabular}}
  \vspace{-0.3cm}
\end{table}

\noindent \textbf{Ave. prediction time.}
Another concern is how early the LC can be recognized, which represents the length of the prediction horizon. Considering that the actual LC duration is about $3.0$ s to $5.0$ s, according to previous research~\cite{toledo2007modelinglanechange,woo2017detectiononprediction}, a clear LC maneuver roughly occurs at $1.5$ to $2.5$ s before crossing the dividing line. The average prediction time of VLSTM is $1.999$ s, which is reasonable if the prediction only relies on the maneuver history of the vehicle itself. The interaction-aware models all improve the average prediction time, which verifies the conjecture that interaction can help to predict behaviors over an extended horizon. Among all the interaction-aware models, our proposed method achieves the largest average prediction time of $2.622$ s, which means that it is better at capturing behaviors in the distant future.

\begin{figure}[t]
  \vspace{+0.4cm}
  \centering
  \begin{subfigure}[b]{0.245\textwidth}
    \includegraphics[width =\textwidth]{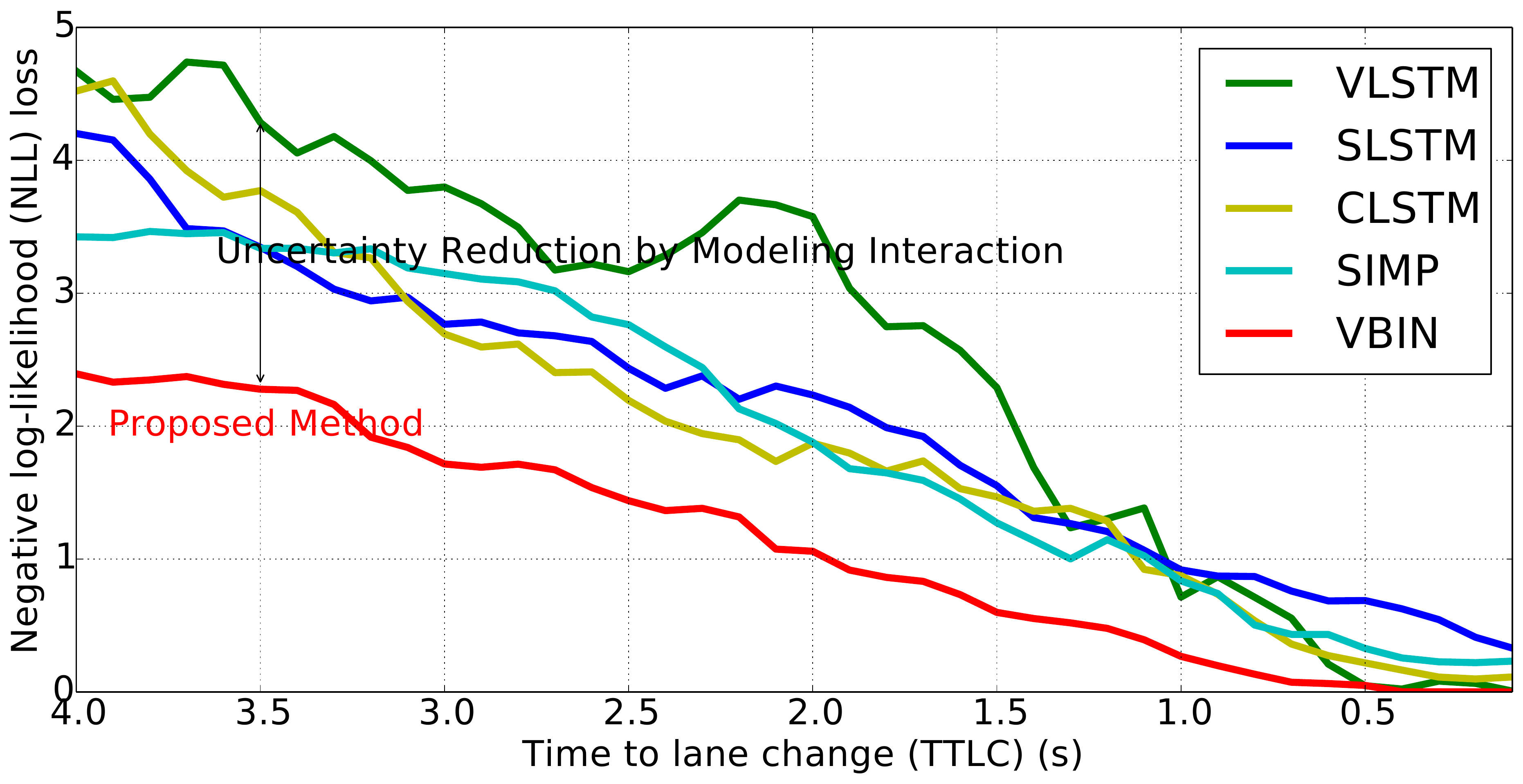}
  \end{subfigure}%
  \begin{subfigure}[b]{0.245\textwidth}
    \includegraphics[width =\textwidth]{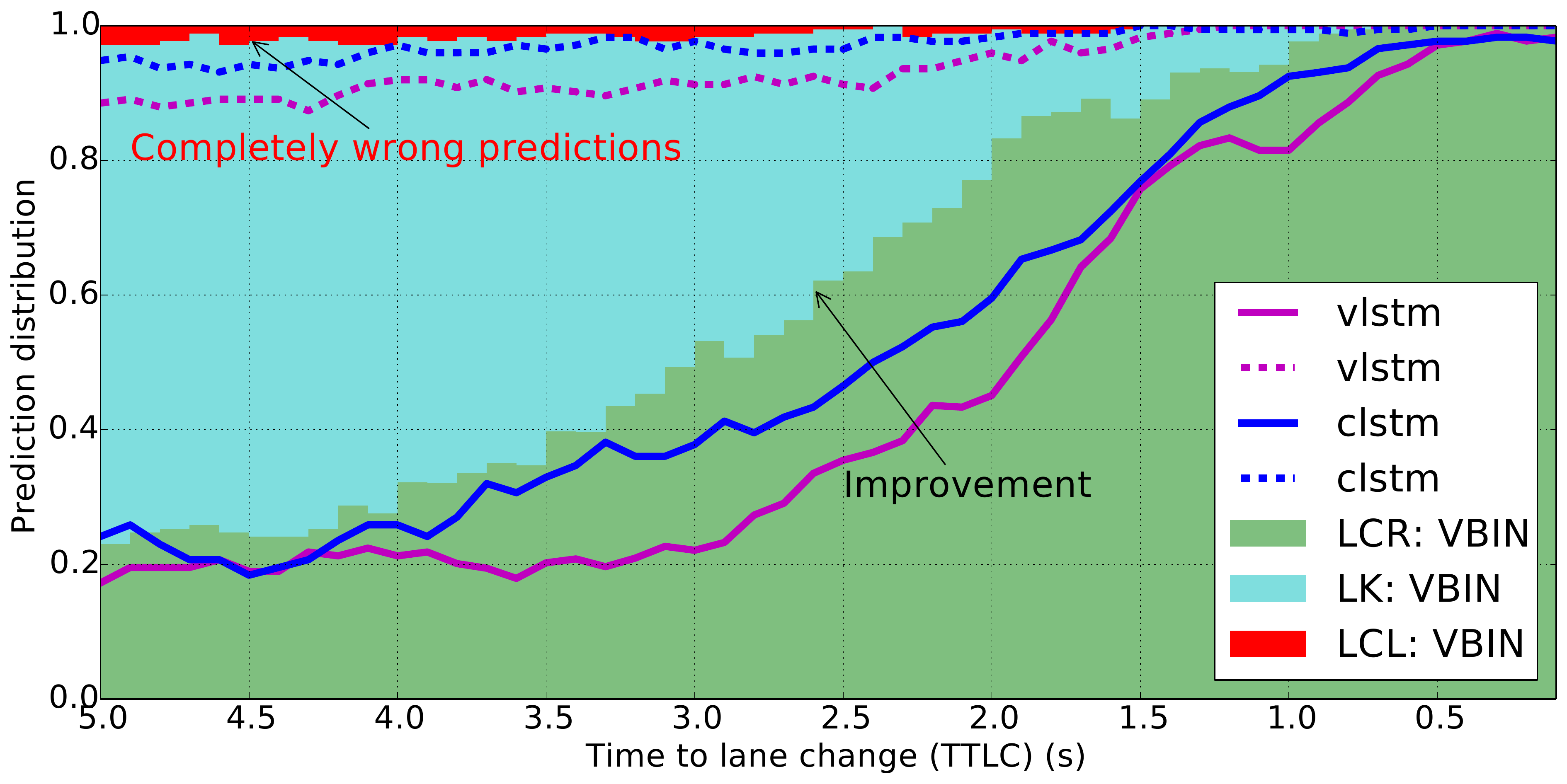}
  \end{subfigure}
  \caption{Illustration of the prediction accuracy w.r.t. the TTLC.} \label{fig:uncertainty_ttlc}
	\vspace{-0.7cm}
\end{figure}

\noindent \textbf{Accuracy with respect to TTLC.} How prediction accuracy changes with respect to time is also a focus of our interest. As shown in Fig.~\ref{fig:uncertainty_ttlc}, we measure the accuracy from two different perspectives: NLL loss and prediction distribution. The NLL loss generally represents the accuracy of the multi-class classification, and our proposed method achieves the lowest NLL loss for all different TTLCs. It can be observed that, compared to VLSTM, significant NLL loss reduction can be achieved, which confirms the usefulness of modeling interaction.

We also illustrate the distribution of the three behavior predictions at different TTLCs for \textit{lane change right} (LCR) cases. The fractions of the three kinds of predictions of our method are stacked for each TTLC. We represent the distribution of VLSTM and CLSTM through lines in Fig.~\ref{fig:uncertainty_ttlc}. It is shown that our method outputs the LCR prediction at the earliest phase, while outputting the least lane change left (LCL) predictions. The distribution also confirms that our proposed method has a superior prediction time.

\noindent \textbf{Social Pooling vs. VBIN.}
In Fig.~\ref{fig:comparison_vlstm}, we provide an example of VLSTM showing the qualitative performance without modeling interaction. In Fig.~\ref{fig:comparison_clstm} and Fig.~\ref{fig:comparison_vbin}, we find that both CLSTM and our VBIN discover the LC intention earlier, and at a TTLC of $2.5$ s, both methods predict LC with a likelihood of over $80$ percent. However, at a TTLC of $3.6$ s, CLSTM cannot capture the LC intention, despite LC being likely, since the target vehicle has a higher velocity than the vehicle in front. Interestingly, in this case, the neighboring vehicle to the right of the target vehicle is actually conducting a rightward LC and leaving more room for the target vehicle. However, it seems that CLSTM fails to capture this information and thinks the neighboring vehicle to the right will block the target vehicle's LC, which matches our conjecture about the drawback of social pooling in Fig.~\ref{fig:occu_pooling}. The proposed VBIN can address this problem and output a reasonable prediction.

\section{Conclusion and Future Work}
\label{sec:conclusion}
In this paper, we present a VBIN for vehicle behavior prediction. The proposed method is capable of predicting vehicle behaviors of an extended horizon by taking vehicle interaction into account. The VBIN is designed for highly dynamic on-road driving and has a novel interaction-aware network structure. It is also scalable and end-to-end trainable. We conduct extensive experiments and compare the proposed method with four baseline methods, including two state-of-the-art methods, showing significant improvements in accuracy and uncertainty reduction.
\clearpage
\bibliography{paper}
\end{document}